\documentclass[letterpaper, 10 pt, conference]{ieeeconf}
\IEEEoverridecommandlockouts 
\usepackage{hyperref}
\hypersetup{
    colorlinks=true,
    linkcolor=blue,
    filecolor=magenta,      
    urlcolor=cyan,
}
\usepackage{amsmath,amsfonts}
\usepackage{array}
\usepackage[font=footnotesize]{caption}
\usepackage{enumerate}
\usepackage[utf8]{inputenc}
\usepackage[T1]{fontenc}
\usepackage{amssymb}
\usepackage{tabularx}
\usepackage{graphicx}
\usepackage{subcaption}  

\usepackage{algorithm}  
\usepackage{algpseudocode}  
\usepackage{breqn}
\usepackage{textcomp}
\usepackage{stfloats}
\usepackage{url}
\usepackage{verbatim}
\usepackage{cite}

\usepackage{color}
\usepackage{colortbl}
\usepackage[most]{tcolorbox}
\let\labelindent\relax
\usepackage{enumitem}
\usepackage{multirow}
\usepackage{diagbox}
\usepackage{booktabs}
\usepackage{xcolor}
\usepackage{listings}

\title{\LARGE \bf Mixed-Agent Museum Tour Guide Design Improves Gendered Learning Outcomes and Visitor Preferences }

\author{Annette M. Masterson$^{4,\dag}$, Wonse Jo$^{5,\dag, *}$, Helena C. Sieh$^{1}$, Lionel P. Robert, Jr.$^{1,4}$, and Dawn Tilbury$^{1,2,3}$%
    \thanks{$^{1}$ Department of Robotics, University of Michigan, Ann Arbor, MI 48109, United States
    }%
    \thanks{$^{2}$ Department of Mechanical Engineering, University of Michigan, Ann Arbor, MI 48109, United States
    }%
    \thanks{$^{3}$ Department of Electrical Engineering and Computer Science, University of Michigan, Ann Arbor, MI 48109, United States
    }%
    \thanks{$^{4}$ School of Information, University of Michigan, Ann Arbor, MI 48109, United States
    }%
    \thanks{$^{5}$
    A+HRI Laboratory, Department of Information and Telecommunication Engineering, Incheon National University, Incheon, South Korea. {\tt\small jow@inu.ac.kr}
    }
    \thanks{$\dag$: equal contribution}%
    \thanks{$*$: corresponding author}%
}

\begin{document}

\maketitle

\begin{abstract}
Robots are increasingly integrated into everyday contexts, including museums, where they can both entertain and educate visitors. To enhance visitor experience and engagement, we present a novel mixed-agent tour guide system that combines a physical robot with a projected virtual agent that actively participates in the tour through conversation and interaction, achieving the interaction richness of two mobile agents from a single platform. We validate the system through a within-subjects study with 30 participants to assess engagement, quality of experience, and learning performance. Participants experienced different conversational styles and agent configurations, and data were collected via surveys, behavioral sensors, and interviews. 
Results showed that engagement and quality of experience remained consistent across conditions. 
Learning performance revealed a significant gender-moderated difference: the mixed-agent conditions improved learning performance for female participants. This suggests that the proposed dyadic conversational style in this paper influenced learning performance differently by gender. Nonetheless, in interviews, participants reported a greater preference for mixed-agent teams regardless of gender, citing interaction as a key factor in their experience. 
\end{abstract}

\section{Introduction} 

Museums combine entertainment and education, but they often rely on static exhibits or audio tours that limit accessibility and engagement \cite{garello2025next}. To enhance experiences, tour guide robots have been proposed as an engaging and financially feasible alternative \cite{garello2025next}. Developing robot tour guides capable of delivering immersive, socially responsive interactions that can be evaluated with comprehensive behavioral and self-report metrics remains an ongoing challenge. In contrast to the dual-robot frameworks discussed by Velentza et al. \cite{velentza2021one}, our design employs a mixed-agent approach that integrates a gimbal-mounted projector with inverse kinematics and path planning, enabling a projected virtual agent to dynamically reposition and actively participate in the tour through conversation alongside the physical robot. This achieves the interaction richness of two mobile agents from a single robotic platform. We validate that this mixed-agent system, especially when paired with different conversational styles, yields superior learning performance for women and increased visitor preference over a single-robot configuration.



Our design replaces the headset requirement of VR tours with a unified robotic system \cite{aoki2023mixed}, offering a scalable solution without the massive investment in individual hardware, mimicking human-guided tours which may not be affordable for all museums \cite{garello2025next}. To determine the value of our system on engagement and learning performance, we conducted a within-subject user experiment with three conditions: 1) a single physical robot with cheerful dialogue; 2) a mixed-agent (physical plus virtual) with storytelling dialogue; and 3) a mixed-agent with bantering dialogue. The scenario represents a robot-guided tour in an indoor environment (e.g., a museum gallery). We measured engagement and learning performance through behavioral, survey, and interview methodologies.




This paper makes several contributions. First, it introduces and evaluates a mixed-agent (physical and virtual) tour-guide system with multiple conversational styles. Second, we provide preliminary evidence of gender-moderated learning benefits for mixed-agent teams, qualitative feedback on the preference for mixed-agent teams, and an observed tension between engagement and learning. These findings advance human-robot interaction within multi-modal systems.

This paper examines how robots and mixed-agent teams influence engagement and learning outcomes. We introduce a multi-application tour guide system with built-in behavioral metrics with the potential for real-time engagement assessment. To conclude, we validate the system’s impact on learning and engagement through a targeted user experiment.


\section{Background} 

\subsection{Entertaining and Learning with Tour Guide Robots}
Static displays and audio museum tours have long been the industry standard; however, autonomous systems offer a level of dynamic interaction that correlates with superior visitor retention and satisfaction compared to traditional, non-adaptive methods \cite{maniscalco2024towards, natori2025impact}. Social signals--facial expressions, movement, and communicative features--consistently increase perceived sociability and likability of museum robots, particularly when designs emphasize personable traits and positive affect \cite{stals2025robot}. While not often addressed in tour guide work \cite{velentza2020museum}, learning can be a subjective or gendered process potentially influenced by social conditioning \cite{onnela2014using}. Considering learning with robot guides, more engaging robots can be viewed more positively \cite{engwall2021robot, wienrich2024gender}. Therefore, demographic differences should be acknowledged in system evaluations to better inform future development.


Recent work integrates tour guide robots \cite{maniscalco2024towards} and augmented reality (AR) agents \cite{gan2023design} to amplify engagement in learning environments. For example, embodied AR agents and emotionally expressive robots outperform audio‑only presentations on measures of preference and perceived expressiveness \cite{velentza2020museum, natori2025impact}. While autonomous or semi-autonomous movements help robots mimic the experience of a human tour guide \cite{rosa2024tour}, their true advantage lies in their lack of constraint. Unlike human guides, robotic systems can seamlessly integrate multi-modal interactions, creating novel and engaging experiences that transcend traditional tours. Still, limited work has been done on \textit{mixed-agent} teams and their impact on engagement and learning performance, particularly with ambulatory robot guides.

\subsection{Mixed-Agent Engagement}


Physical robots can enhance engagement in informal learning environments \cite{natori2025impact, velentza2020museum}. Similarly, virtual robots, either projected onto the wall \cite{ro2019projection}, 
as an augmented reality agent \cite{lee2017application}, 
or through a mobile application \cite{pliasa2021interaction}, have been found to increase learning performance. 

In response to enhanced learning and engagement, we designed a tour experience with a combination of agents and integrating multi-modal interactions to expand tour guide offerings. Integrating augmented reality bots that can manipulate video clips with a physical robot allows for more effective interactions with participants \cite{han2010museum}. Additional work reports that mixed configurations can sustain engagement \cite{velentza2021one, chen2022robot} and, in some cases, improve learning performance by distributing content across agents \cite{chen2022robot}. However, Velentza et al. \cite{velentza2021one, velentza2019human} found a complex relationship between robot configurations and learning performance. A single-robot setup improved knowledge recall, but a two-robot setup led to lower retention of the tour content, as assessed in the post-experiment quiz. Curiously, despite the diminished learning, participants in the two-robot condition rated their experience as more pleasant. This highlights the importance of testing learning performance across different robot configurations to balance learning performance \cite{velentza2022human}, particularly with a novel \textit{mixed-agent} system. In this study, we propose the following hypotheses to guide our design and evaluation:



\textit{\textbf{H1:} Mixed-agent teams will have increased quality of experience, engagement, and learning performance over a single robot}. 

\textit{\textbf{H2:} Mixed-agent teams bantering conversational style will have higher quality of experience, engagement, and learning performance over the storytelling style}.

\section{Mixed-Agent Tour Guide System}
\begin{figure}[t]
    \centering
    \vspace{2.5mm}
    \includegraphics[width=1\linewidth]{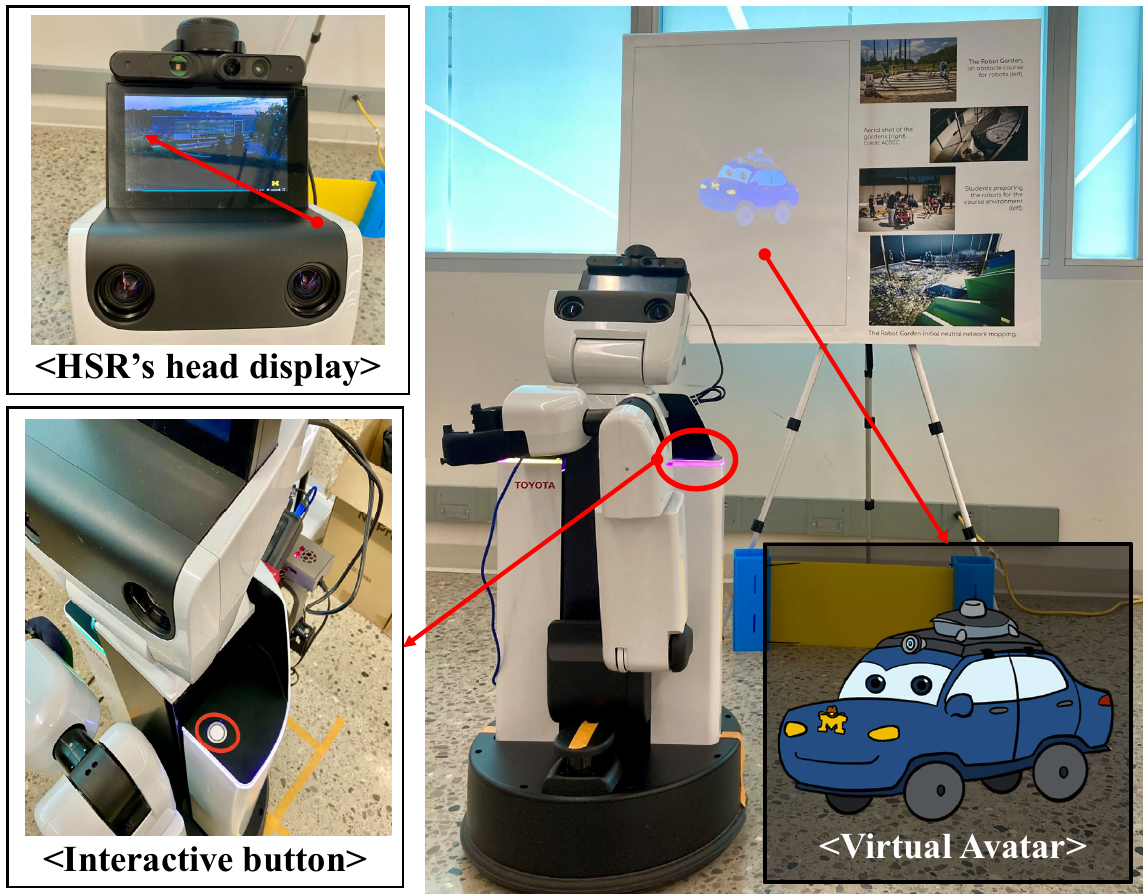}
    \captionsetup{width=1\linewidth}
    \caption{Details of the robot platform, including the robot's head display and forward-facing depth camera (top left), the interactive shoulder button (bottom right), the rear-facing RealSense Depth Camera (D345i, bottom left), and examples of projections (e.g., image, video, and our virtual avatar).}
    \label{fig:robot_projectors}
\end{figure}

\begin{figure}[t]
    \centering
    \includegraphics[width=1\linewidth]{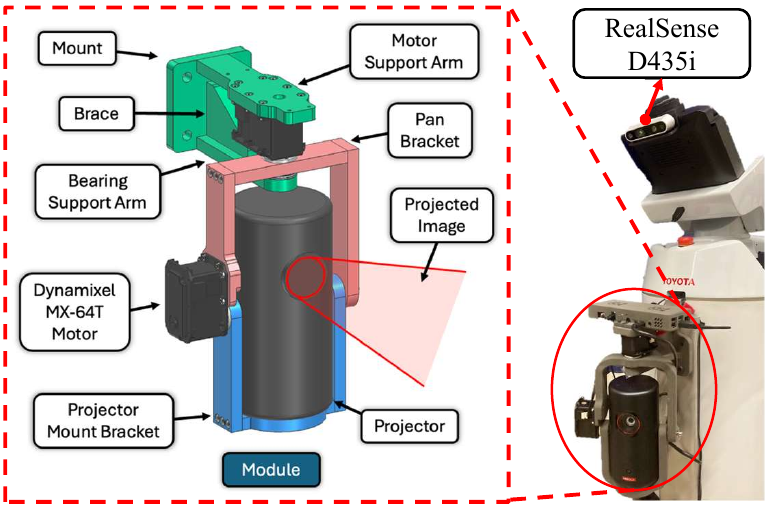}
    \captionsetup{width=1\linewidth}
    \caption{Customized projector to augment the virtual avatar.}
    \label{fig:robot_projectors_details}
\end{figure}

As illustrated in Fig.~\ref{fig:robot_projectors}, we propose a novel mixed-agent team comprising a physical service robot platform (Toyota HSR) and a projected virtual avatar generated by a laser projector mounted on the HSR's back (see Fig.~\ref{fig:robot_projectors_details}). The system's key technical contributions include: (1) multi-location virtual agent 
positioning through gimbal-mounted projection, enabling the virtual agent to appear at exhibit locations, (2) synchronized multi-agent coordination across processors (desktop RTX 4090, Jetson Nano, Raspberry Pi 4) to coordinate speech and movements of both agents, and (3) integrated behavioral measurements embedded within the robot's control loop, enabling continuous data collection (reaction times, physical distance, head orientation). The overall system of the mixed-agent team communicates via the Robot Operating System (ROS) \cite{Quigley09}, which enables data sharing across a distributed system. 
The system includes three key ROS packages: Tour Script, which enables a mixed-agent team to explain the exhibits; Navigation, which drives to predefined exhibit locations; and Engagement Tracking, which measures participants' distance, head pose, and reaction time in response to the robot's prompts. Distinct voices were chosen for the physical and virtual agents \cite{openai2024tts}. To ensure audio consistency across participants, generated audio files were then downloaded and uploaded directly to the robot's system.

\subsection{Physical Robot Platform (HSR)}
The HSR is equipped with the Navigation package that uses 2D Lidar and two head-depth cameras (front and rear views), enabling a safe navigation system that automatically avoids obstacles. We implemented the engagement tracking package on the HSR that maintains eye contact with participants throughout the tour \cite{kompatsiari2019measuring}. 
The package includes PID-controlled pan/tilt motors on the HSR's head continuously track helmet-mounted ArUco markers worn by participants, keeping them centered in the head camera frame for face-to-face orientation. To robustly and directly estimate the 6D pose of the human participant's head during the interaction, we utilized ArUco fiducial markers affixed to the safety helmet rather than relying on markerless vision-based frameworks like MediaPipe. This tracking capability not only enables us to measure the distance between the participant and the HSR, but also to calculate the head reaction time (RT) based on the robot prompts during the tour. 

\subsection{Virtual Avatar}
The virtual avatar (VA) is designed as a cartoon-style vehicle with anthropomorphic features, a design choice intended to enhance user engagement \cite{maniscalco2024towards} and mitigate potential discomfort \cite{dautzenberg2024follow}. We chose not to include a humanoid virtual agent in order to better distinguish the physical and virtual entities as distinct agents. The VA is projected via a gimbal-mounted projector, actuated by two servo motors providing a pan range of $\pm55^\circ$ and a tilt range from $-90^\circ$ to $+45^\circ$. To ensure precise visual delivery, projection locations are pre-programmed relative to the robot’s pose at each tour stop. Gimbal orientations are computed using inverse kinematics, while smooth motion trajectories are planned via convex quadratic programming.

The VA’s animations are rendered using the PyGame framework, supporting frame-by-frame playback of animated sprites at a refresh rate of approximately 15 Hz. For realistic interaction, the avatar’s mouth movements are synchronized in real-time with OpenAI’s text-to-speech engine \cite{openai2024tts}. In addition to the avatar, the projector serves as a versatile display for supplementary visual media, such as images and videos, which are projected onto walls during exhibit explanations and onto the floor to assist with navigation.

\section{Validation Experiment}

\begin{figure*}[!ht]
    \centering
    \vspace{2.5mm}
    \begin{subfigure}{0.8\linewidth}
        \centering
        \includegraphics[width=\linewidth, alt = {Image of open room with six exhibit posters showing the directions of the experiment}]{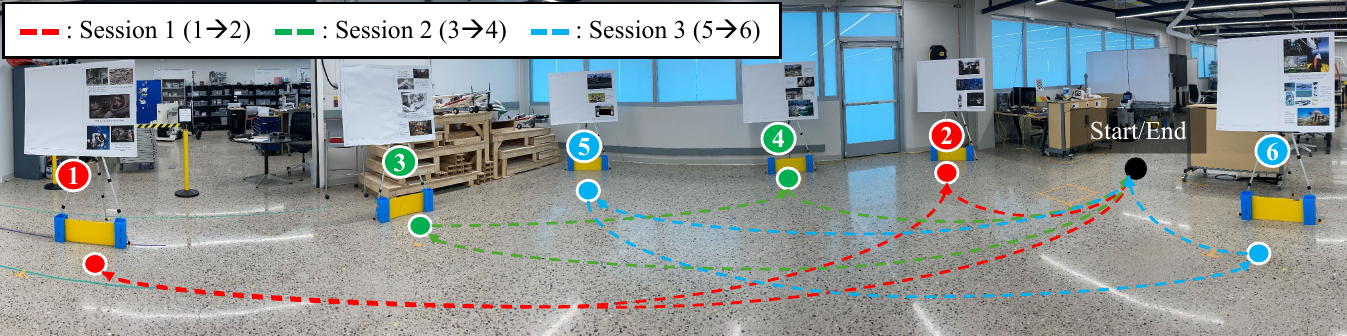}
        \caption{}
        \label{fig:exper_layout}
    \end{subfigure}      
    \hspace{0pt}
    \begin{subfigure}{0.18\linewidth}
        \centering
        \includegraphics[width=\linewidth, alt = {Fig.~detailing the experimental procedure going from the start through the introduction and conditions to the survey, interview, and the end}]{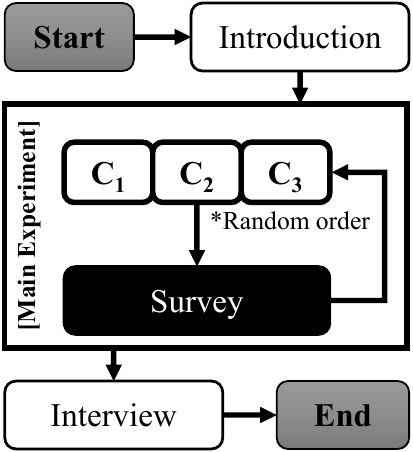}
        \caption{}
        \label{fig:exp_protocol}
    \end{subfigure}
    \caption{Overview of the experimental setup and protocol: (a) Experiment environment and (b) Procedures for the user experiment, in which the order of the condition ($C$) was randomized.}
    \label{fig:exp_combined}
\end{figure*}

To validate the efficacy of the proposed mixed-agent team, we conducted a user study focusing on how this configuration influences participants' engagement, quality of experience, and learning performance. During the experiment, subjects participated in a robot-led museum tour where the mixed-agent team delivered interactive presentations about various exhibits. This study was conducted with the approval of the Institutional Review Board (\#IRB-HUM00278018). 

\subsection{Participants}
For this user experiment, we recruited 30 participants (14 female and 16 male; mean age = 32, SD = 13) based on a post-hoc power analysis conducted using G\*Power \cite{faul2007g}, with an effect size of \textit{f} = 0.24, a significance level ($\alpha$) of $.05$, and a power of $.80$. 
All participants confirmed their eligibility to participate in the study as follows: (1) 18 years old, (2) have normal or corrected vision and hearing, and (3) have the ability to follow the robot for an hour. 

\subsection{Study Design} 
The experiment employed a within-subjects design comprising three conditions (60 minutes total), presented in a random order to minimize practice effects \cite{jo2024cognitive}. The conditions differ based on the type of information projected by the robot’s projector or the style of conversation within the mixed-agent team. Factual verbal (information about the exhibits) and visual (videos and images related to the exhibits) content remained identical across all conditions. There are three conditions: $C_{1}$) a single physical robot with cheerful dialogue; $C_{2}$) a mixed-agent team featuring storytelling conversational style, with the agents not engaging with each other; and $C_{3}$) a mixed-agent team interacting humorously in bantering conversation style with each other. The baseline consisted of a single robot speaking in a cheerful manner; Condition 1 presented the same videos as the other conditions but was not accompanied by a virtual agent. 

Experiments were conducted in a museum-like environment (see Fig.~\ref{fig:exper_layout}) with participants wearing a helmet-mounted camera for tracking their behavior. Exhibits for this study included six distinct academic posters: departmental history, bipedal robotics, collaborations, robot training facilities, autonomous vehicles, and aquatic integration. 

The scripts in each exhibit were written by researchers and then fed into ChatGPT-4 to generate ideas for cheerful and humorous styles, designated the storytelling--without dialogue between agents--and bantering--humorous dialogue between agents--conditions, respectively. The purpose of the stylistic change is to determine whether communication style helps improve engagement and learning performance. Researchers compiled scripts to align with the conditions, then presented them in randomized order. As an example, scripts for the UGVs exhibit included: Single HSR robot: “\textit{The robot was being trained as a UGV — that means Unmanned Ground Vehicle. No driver needed}.” Storytelling: HSR robot– “\textit{The robot was being trained as a UGV, that means Unmanned Ground Vehicle. No driver needed}.” virtual agent– “\textit{I’m Scout, a UGV, but I’m way more updated than these bots in 1992.}” Bantering: HSR robot–“\textit{The robot was being trained as a UGV — that means Unmanned Ground Vehicle. No driver needed. Wait a minute, are you a UGV, Scout?}” virtual agent– “\textit{That’s right, Remy! But I’m way more updated than these bots in 1992.}”

\subsection{Survey Procedures}

Fig.~\ref{fig:exp_protocol} details the overall procedure used in this study. The introduction survey included the pre-quiz for learning performance and demographic questions: age, gender, ethnicity, and education. The introduction survey also asked participants about their prior experience and familiarity with service robots. Participants completed two questionnaires to assess their engagement and quality of experience at the end of each condition: 1) User Engagement Short Scale (UES) \cite{o2018practical}, and 2) Quality of Experience (QoE) scale \cite{rosa2024tour} to assess satisfaction. 
In addition to the self-reported questionnaires, Learning Performance (LP) was measured by comparing pre-experiment quiz scores with post-session quiz scores at the end of each condition, using consistent questions and answers across conditions. Quizzes were written by researchers from the information in the scripts. Quiz questions were inputted into ChatGPT-4 to evaluate the complexity. Quizzes were finally assessed during pilot tests to balance difficulty and achievability. Quiz answers were multiple choice and were answered in the exhibits. An example of a quiz question is "What is the maximum speed attainable by the bi-pedal treadmill?"

\subsection{Behavioral Measures} 

\begin{figure}[t]
    \centering
    \vspace{2.5mm}
    \includegraphics[width=0.85\linewidth]{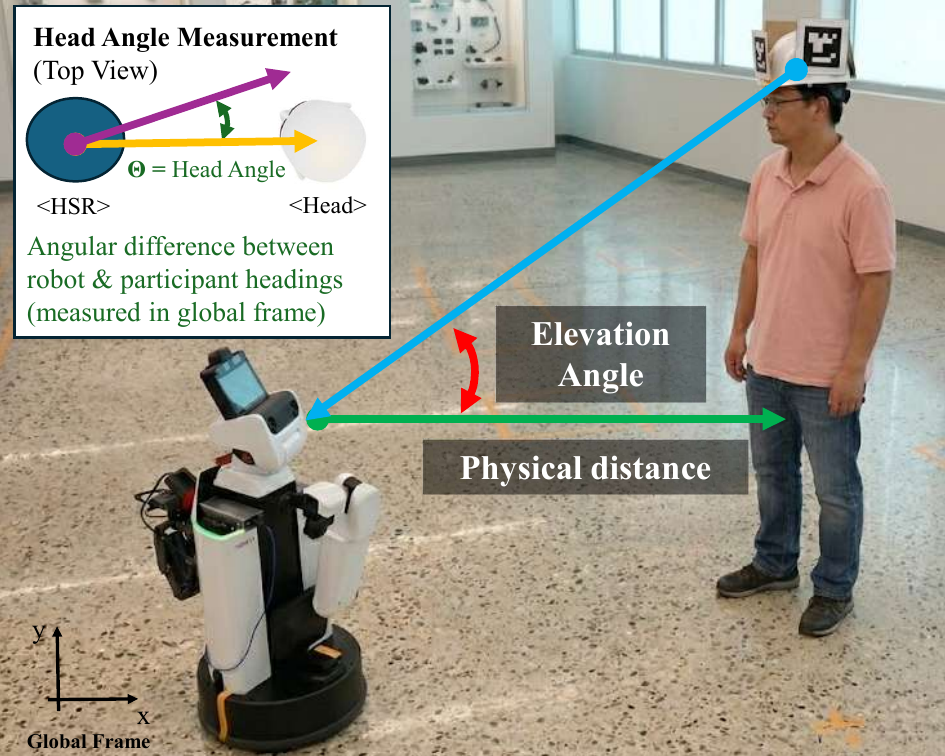}
    \caption{Behavioral measurement setup. ArUco marker tracking provides: (1) physical distance (green line, meters), (2) elevation angle (red arc, degrees), and (3) head angle – orientation difference between robot heading (yellow) and participant heading (purple) (green arc, $\theta$ in degrees).}
    \label{fig:behavior_measure_fig}
\end{figure}

Our mixed-agent system enables real-time tracking of participant behavior throughout the tour (Fig.~\ref{fig:robot_projectors}). As illustrated in Fig.~\ref{fig:behavior_measure_fig}, the behavioral data included: (1) Physical distances -- mean distance between participant and robot measured in meters, indicating spatial comfort and physical distance preferences; (2) Cognitive reaction time (RT) -- mean response latency in seconds from robot's prompt to participant's actual actions, measured four times during experiments; and (3) Head angle -- angular difference between participant and robot heading directions (degrees), measured in the global frame.

For the physical dimensions, the robot uses both front and rear cameras mounted on its head to measure the distance and angle between the robot's location and the participant's head location. This physical distance is not linked to gaze \cite{koller2023robotic}, but physical distance. Once the robot detects the ArUco markers attached to the participant's helmet, it estimates their position in the global map and calculates (1) the distance from the robot to the participant and (2) the bearing angle from the robot's forward direction to the participant's location.

We also recorded two cognitive RT metrics: (1) Head RT: the time it takes for the participant to turn their head to a specific poster after the robot gives a directional prompt and (2) Button RT: the time between the button prompt and the participant's push of the button to initiate video playback on the robot's head-mounted display.

\subsection{Interview Procedures}

To gather in-depth data, the experiment concluded with brief semi-structured interviews. Questions were designed to align with the quantitative scales, such as asking about the most interesting components of the tour to relate to the UES and about the tour experience to relate to QoE. This provided context for the survey and behavioral data. 
All 30 participants completed the interview, with sessions ranging from 4--14 minutes and an average of 7 minutes. We used randomly generated four-digit participant identification numbers (e.g., P1234) for anonymity. Interviews transcribed using Otter.AI, and transcripts were reviewed for accuracy. 

Data analysis was conducted inductively to allow themes to emerge directly from the data by assessing each interview comment as a data point \cite{corbinstrauss2008, patton2001}. Codes were developed through a reflective thematic analysis, an iterative process that identifies themes across the aggregate data \cite{braun2019reflecting, braun2006using, denzinlincoln2018}. Through the data analysis, significant themes emerged focused on knowledge development, robotic capabilities and design, interaction, and mixed-agent preference.

\section{Results}
Our user study ($N=30$) revealed that our mixed-agent team with a bantering conversational style can enhance learning performance for women, and qualitative results illustrate visitor preference across participants. We also reveal a negative correlation between engagement and learning performance. Behavioral results demonstrate women stand closer to the robot, particularly during the bantering condition. Cognitive reaction times illustrate high compliance with prompts.

\subsection{Survey Results} 
We evaluated three measures: (1) Sum of Engagement (UES) scores, (2) Mean of Quality of Experience (QoE) scores, and (3) Learning Performance (LP; post–pre quiz difference normalized to a 0--1 range). Repeated-measures ANOVA (rmANOVA) was conducted via Pingouin \cite{vallat2018pingouin}. 

From the rmANOVA results, we found a significant effect of condition on LP $(F(2,58)=3.95, p=0.025, \eta_p^2=0.12)$, where $C_3$ ($M=0.77, SD=0.25$) outperformed $C_1$ ($M=0.61, SD=0.31$). Post-hoc pairwise comparisons using t-tests revealed that participants in $C_3$ achieved significantly higher learning scores than $C_1$, with a mean difference of $0.16$ ($t(29)=-2.52, p=0.018$). While the difference between $C_1$ and $C_2$ ($M=0.75, SD=0.25$) was marginal ($t(29)=-2.04, p=0.051$), no significant difference was observed between $C_2$ and $C_3$ ($t(29)=-0.32, p=0.752$).
However, no significant differences were found for Sum UES $(p=0.479)$ and Mean of QoE scores $(p=0.614)$, indicating that mixed-agent teams improved overall learning performance, but not perceived engagement or quality of experience.

\begin{figure}[t]
    \centering
    \vspace{2.5mm}
    \includegraphics[width=1\linewidth, alt = {Graph of the gender interaction effect between conditions.}]{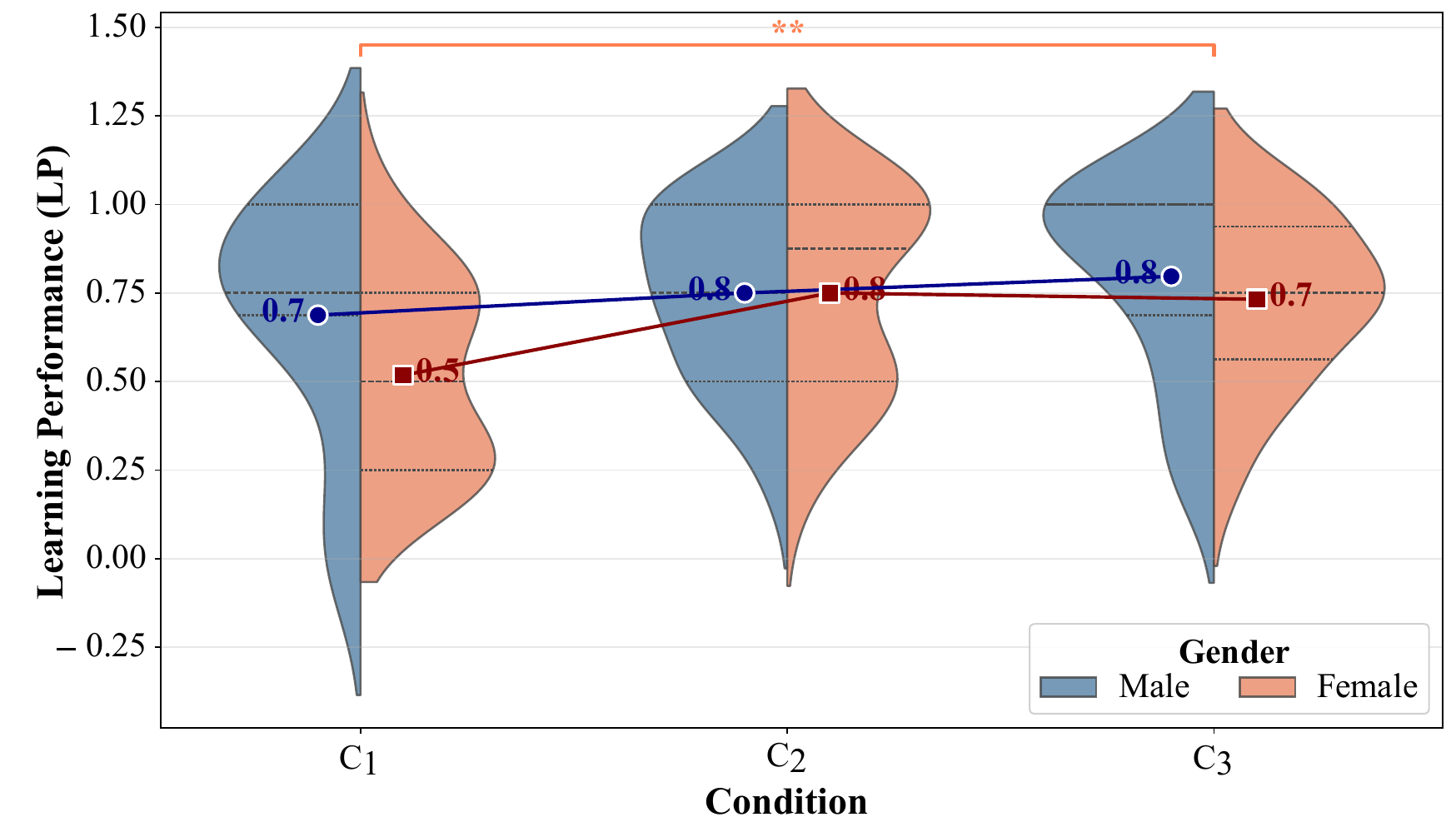}
    \caption{Gender comparison of normalized Learning Performance (LP) across conditions using split violin plots, which is post--pre quiz difference normalized to a 0--1 range. Markers indicate group means; brackets with asterisks show significant pairwise differences ($**: p < 0.05$) for female participants between C3 (bantering, mean = 0.77) and C1 (single robot, mean = 0.5). No significant differences for males.}
    
    \label{fig:stat_results}
\end{figure}

To examine gender differences, participants were divided into 16 males and 14 females, and rmANOVA was conducted on each group. A strong condition effect emerged for LP in the female group $(F(2,26)=3.86, p=0.034, \eta_p^2=0.23)$, where $C_3$ outperformed $C_1$ by a mean difference of $0.21$ ($p = 0.028$), as illustrated in Fig.~\ref{fig:stat_results}.

Therefore, we can conclude that the mixed-agent team outperformed the single robot, primarily in learning performance. While Mean QoE showed a positive numerical trend without reaching statistical significance, gender-specific benefits emerged: males showed a preference for the storytelling dialogue ($C_2$) for experience, while females achieved significantly higher learning gains in the bantering dialogue ($C_3$). This suggests that while mixed agents generally enhance learning, dialogue effectiveness may be gender-dependent.


\subsection{Behavioral Results}
Behavioral analyses included physical distance, head angle, and RT for button presses and head rotations, as prompted by the robot. Participants demonstrated high compliance with interactive prompts: button presses achieved 100\% response rate, while head prompt achieved 68\% response rate, RT showed no significant effects between conditions.

\begin{figure}
    \centering
    \includegraphics[width=1\linewidth, alt = {Fig.~of gender differences (male and female) across physical distance, button reaction time (RT), and head (RT)}]{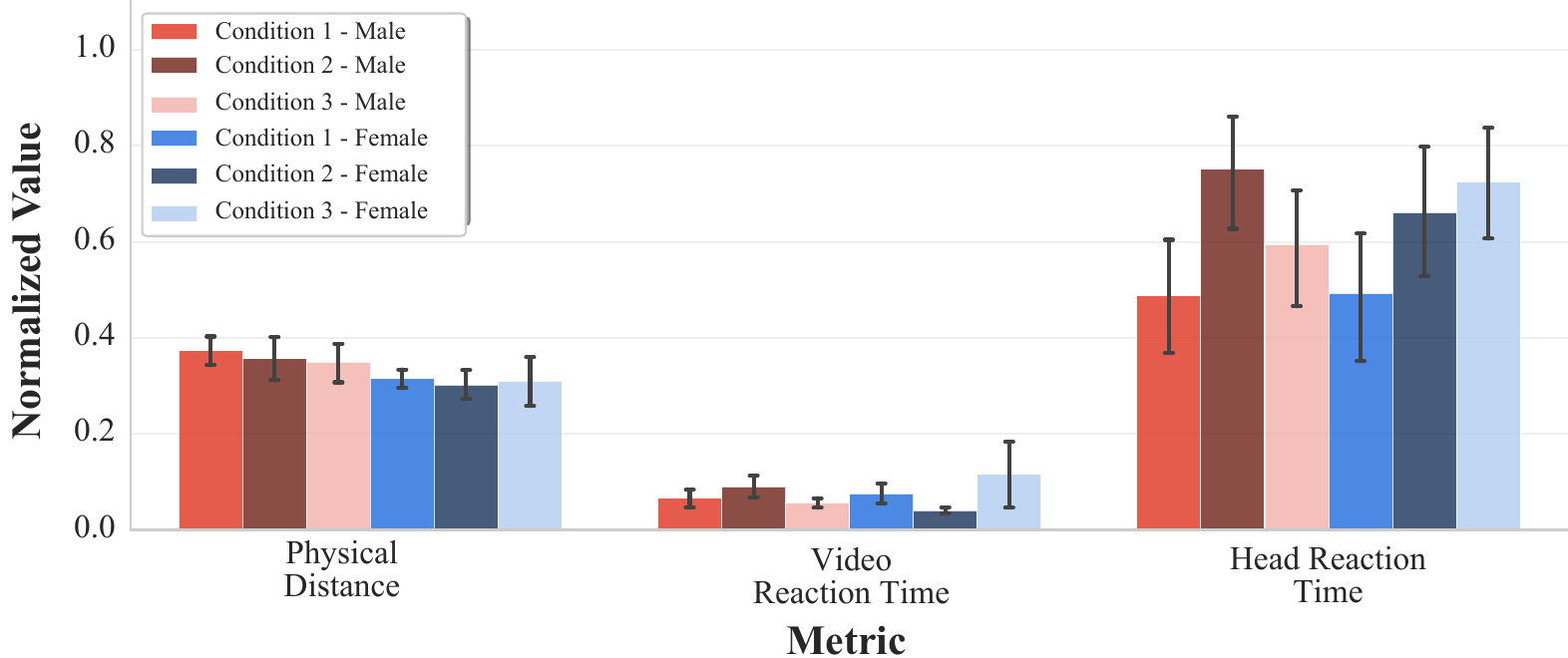}
    \caption{Behavioral measures by gender across conditions. Error bars represent standard error of the mean (N=16 males, 14 females). Males maintained significantly greater physical distance across all conditions (p < .001).
    }
    \label{fig:seaborn_normalized_metric_gender_chart}
\end{figure}

As shown in Fig.~\ref{fig:seaborn_normalized_metric_gender_chart}, male participants consistently maintained greater distances from the robot than female participants across all conditions. Participants stood closest to the robot during $C_{3}$ (bantering) and maintained greatest distance during $C_{2}$ (storytelling), suggesting the interactive bantering style encouraged closer physical engagement. Post-hoc pairwise comparisons revealed significant distance differences between all conditions: $C_{2}$ vs. $C_{1}$ ($MD = 0.0285$, $p < .001$), $C_{2}$ vs. $C_{3}$ ($MD = 0.0161$, $p < .001$), and $C_{1}$ vs. $C_{3}$ ($MD = 0.0124$, $p < .001$). 

\begin{figure}[t]
    \centering
    \vspace{2.5mm}
    \includegraphics[width=1\linewidth, alt = {Fig.~of correlation heatmap with seven statistically significant items.}]{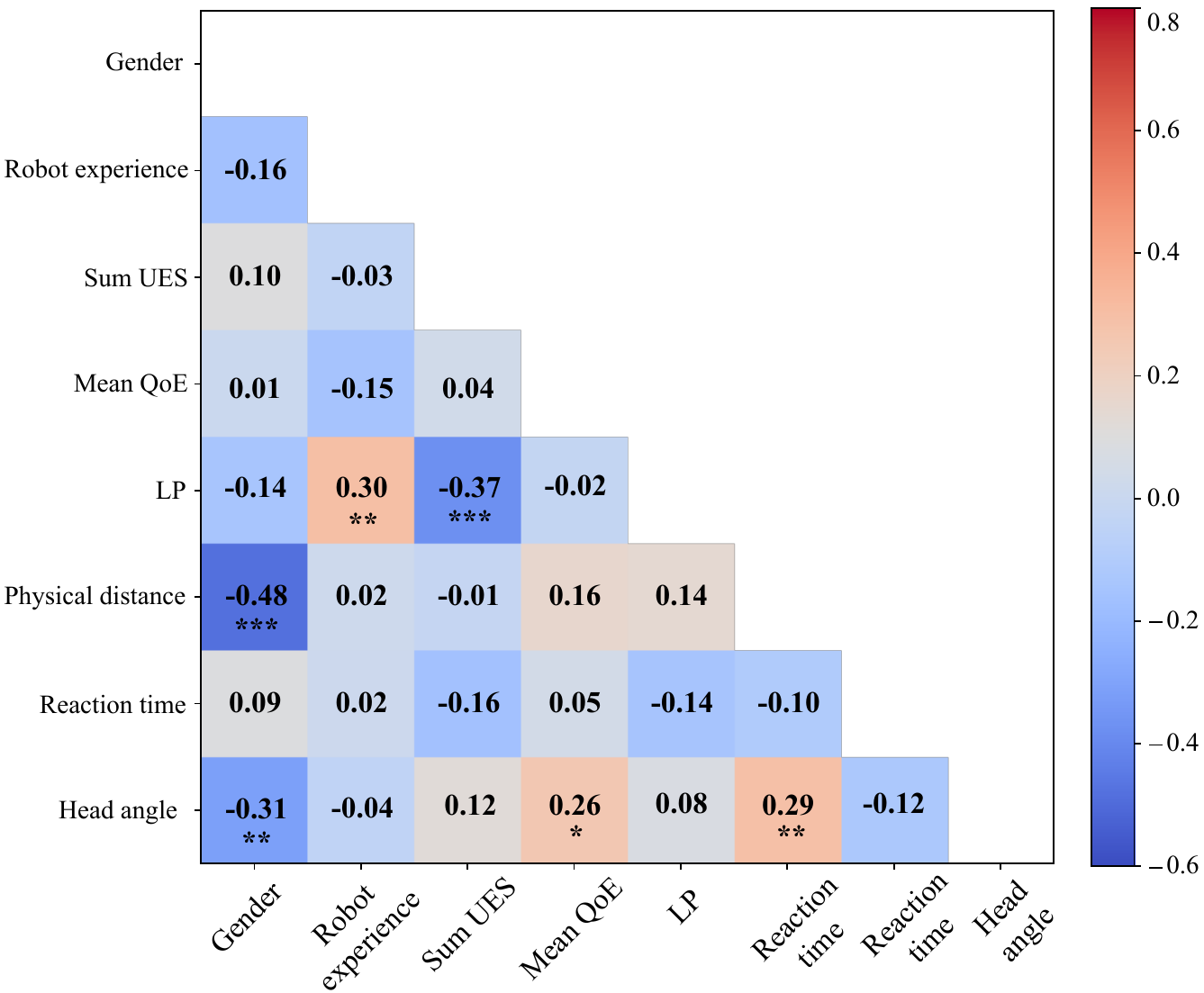}
    \vspace{-15pt}
    \caption{Heatmap of Pearson correlation coefficients among study variables. The values in each cell represent correlation coefficients ($\gamma$) with significance indicated by asterisks ($^{*}: p < 0.05$, $^{**}: p < 0.01$,  $^{***}: p < 0.001$).}
    \label{fig:correlation_heatmap}
\end{figure}

As shown in Fig.~\ref{fig:correlation_heatmap}, gender was negatively correlated with head angle ($r = -0.31$, $p = .003$), suggesting that male participants tended to exhibit larger directional changes relative to the robot than female participants. Head angle was also moderately positively correlated with Mean QoE ($r = 0.26$, $p = .014$), suggesting that participants who engaged in greater head movement reported higher perceived experience quality. 
In addition, physical distance was positively correlated with head angle ($r = 0.29$, $p = .005$), indicating coordinated spatial and directional adjustment during interaction. 

\subsection{Correlation Results} 

We examined correlations between demographic, survey, and behavioral variables using Pearson correlation analyses \cite{vallat2018pingouin}. Demographic variables included gender (1 = male, 2 = female) and robot experience (0 = no prior experience, 1 = prior experience). Survey measures comprised Sum UES, Mean QoE, and LP (post–pre quiz difference).

Several statistically significant correlations of moderate or greater magnitude ($|r| \geq .25$, $p < .05$) were identified. Gender was negatively correlated with physical distance ($r = -0.48$, $p < .001$) and head angle ($r = -0.31$, $p = .003$), indicating that male participants tended to maintain greater distance from the robot and exhibit larger directional shifts. LP was negatively associated with UES ($r = -0.37$, $p < .001$), whereas robot experience was positively associated with LP ($r = 0.30$, $p = .004$). Mean QoE demonstrated a moderate association with head angle ($r = 0.26$, $p = .014$). Physical distance was also positively correlated with head angle ($r = 0.29$, $p = .005$), suggesting coordinated spatial and directional adjustments. 

\subsection{Interview Results} 

Interviews provided deeper context for the quantitative findings, confirming a  preference for mixed-agents (17 out of 29 participants favored the team, with one neutral). Participants noted that the collaborative dynamic—characterized by jokes (P7273) and dialogue (P6342)--emulated human-to-human communication, making the experience feel "\textit{closer to a human being than a robot}" (P8237). This multi-modal delivery provided "\textit{fresh perspectives}" (P7198) by shifting attention between agents, a feature deemed essential in today’s "\textit{attention economy}" (P3470). The ability to direct their attention to the robot's screen between exhibit explanations was also an interactive highlight, with participants enjoying the \textit{"multiple ways that it's showing"} information (P6496), helped them answer the quiz (P1025). 

Regarding gender, while no explicit preference was found, female participants particularly highlighted the "\textit{creative}" and "\textit{cute}" aspects of the team (P5086), noting that the virtual avatar's animated eyes made interaction "\textit{easier}" and more engaging (P2990). Female participants found the \textit{"talking"} (P5781) and \textit{"interaction"} (P7198) between the agents increased their engagement with the content.


Two primary design challenges emerged: speed and navigation. Although the robot’s speed ($0.8$ km/h) ensured a sense of security (P8871), participants found it significantly slower than a natural human pace, forcing them to moderate their pace (P7601). These findings suggest that future iterations must balance cautious, safe navigation with a more natural walking speed to enhance real-world viability.

\section{Discussion}


\subsection{Summary of Results}
Perceptions of engagement and quality of experience remained consistent across conditions in the survey results. Hypothesis 1 was partially supported with learning performance increasing for women in mixed-agent teams. Hypothesis 2 partially supported as the bantering style increased learning performance for women. We also found a negative correlation between engagement and learning performance, consistent with prior work \cite{velentza2020museum}, indicating that greater perceived engagement does not necessarily translate to improved learning even in real-world settings. The qualitative interview data offered deeper insights, with participants reporting a strong preference for mixed-agent systems. This favor was attributed to the robot's multi-modal capabilities; features like the projector and shoulder button transformed the interaction that felt more dynamic than traditional setups.

Behavioral data revealed significance in head angle and physical distance. Head angle showed no significant condition effects but was significantly correlated with gender, QoE, and reaction time. Notably, females tended to stand closer to the robot than males, contradicting previous research \cite{takayama2009influences} potentially due to contextual differences in robotic appearance \cite{saeki2024sequential} or politeness \cite{zojaji2025impact}. 

\subsection{Mixed-Agents and Learning Performance}
Our results show that mixed-agent effectiveness is highly demographic-dependent; specifically, female participants achieved higher learning performance. This may be due to an overarching preference for talkative robots, particularly when handling more tasks \cite{wienrich2024gender}, as collaborative environments may be more aligned with women's interaction style \cite{onnela2014using}. Women also tend to rate robots that are more personable and have higher attention switches as better for learning \cite{engwall2021robot}. We believe our mixed-agent system may mimic previous research attuned to women's learning styles and preference for a more collaborative, talkative, and personable robot experience. However, additional analyses should be conducted to reaffirm these findings on conversational style and mixed-agent preferences. Our findings suggest a need for personalized conversational styles to match gendered preferences. While the bantering style did not universally improve all metrics, qualitative data confirmed that the mixed-agent configuration offers a uniquely engaging experience beyond other traditional tours \cite{natori2025impact}.

Thus, this study validates the trade-off between visitor enjoyment and learning performance in a real-world context. Although dual-agent setups can hinder information retention while increasing pleasure \cite{velentza2020museum}, our gender findings and interview data, in particular, demonstrate that a properly balanced mixed-agent system can enhance both simultaneously. While guided tours reduce cognitive load compared to audio tours \cite{van2012effect}, the mental demand remains a challenge in robot-led interactions and must be addressed. Managing this tension requires a dynamic interaction design that recalibrates its strategy throughout the tour to maintain an optimal balance between entertainment and education.

\subsection{Implications for Design and Implementation}
Our design enhancements offer three key implications for social robotics: (1) Gendered differences necessitate personalized, inclusive content \cite{stals2025robot}. Our mixed-agent and behavioral metric system can become a platform for real-time adjustments to content. Systems should utilize real-time metrics—like reaction times—to dynamically adjust information flow, using learning checks during high-engagement segments to mitigate cognitive overload. Removing the dependency on ArUco markers would allow behavioral metrics to be integrated more naturally into real-world environments; (2) Multi-modal features should align with user expectations to avoid the "uncanny valley" \cite{mori2012uncanny} or an "expectation gap" \cite{kwon2016human}. Designers should harmonize agent capabilities, expressivity, and timing to reduce dissonance; (3) Robot navigation must mimic human social conventions. While $0.8$ km/h provides safety, it forces users to unnaturally moderate their pace. Future designs should integrate transparent safety cues \cite{tamai2022guiding} while striving for a more natural walking speed that maintains the fluidity of human guidance.

\section{Conclusion and Future Work}
We developed a novel mixed-agent museum tour guide system that achieves the interaction richness of mixed-agents through a single robotic platform and a gimbal-mounted projector. Our user study ($N=30$) revealed that this multi-modal configuration significantly enhances visitor preference and improves learning performance, particularly for women, while identifying a subtle trade-off between engagement and knowledge retention. Although the study was limited by its controlled lab setting and the subjective nature of humor \cite{martin2018psychology}, it underscores the value of dynamic, ambulatory robot teams. 

Future work should focus on: (1) integrating LLM-based personalization, allowing for choice in mixed-agent and personality styles, (2) balancing engagement with learning performance by testing cognitive load, and (3) validating the mixed-agent system’s navigation in real-world, high-traffic museum environments to accommodate diverse user demographics and abilities. Future research should leverage these real-time behavioral metrics to further inform tour guide systems, creating a more robust framework for quantifying engagement with objective data across increasingly complex social environments.

\section*{Acknowledgment} 
This work was supported in part by funding from the University of Michigan (U-M), U-M's Undergraduate Research Opportunity Program (UROP), and U-M's undergraduate capstone students (ME450 and ROB450). We thank them for their invaluable skills and time in developing the mixed-agent tour guide system.

\typeout{}
\bibliography{main}
\bibliographystyle{IEEEtran}
\end{document}